# Deep Neural Networks Ensemble for Detecting Medication Mentions in Tweets

Davy Weissenbacher 1, Abeed Sarker 1, Ari Klein 1, Karen O'Connor 1, Arjun Magge Ranganatha 2, Graciela Gonzalez-Hernandez 1

1. Department of Biostatistics, Epidemiology and Informatics, Perelman School of Medicine, University of Pennsylvania, Philadelphia, PA 19104, USA
2. Biodesign Center for Environmental Health Engineering, Biodesign Institute, Arizona State University, Tempe, AZ 85281, USA

Corresponding author: Davy Weissenbacher, 480-492-0477, 404 Blockley Hall, 423 Guardian Drive, Philadelphia, PA 19104-6021, dweissen@pennmedicine.upenn.edu

## Abstract

**Objective** After years of research, Twitter posts are now recognized as an important source of patient-generated data, providing unique insights into population health. A fundamental step to incorporating Twitter data in pharmacoepidemiological research is to automatically recognize medication mentions in tweets. Given that lexical searches for medication names may fail due to misspellings or ambiguity with common words, we propose a more advanced method to recognize them.

**Methods** We present *Kusuri*, an Ensemble Learning classifier, able to identify tweets mentioning drug products and dietary supplements. *Kusuri* (薬, "medication" in Japanese) is composed of two modules. First, four different classifiers (lexicon-based, spelling-variant-based, pattern-based and one based on a weakly-trained neural network) are applied in parallel to discover tweets potentially containing medication names. Second, an ensemble of deep neural networks encoding morphological, semantical and long-range dependencies of important words in the tweets discovered is used to make the final decision.

**Results** On a balanced (50-50) corpus of 15,005 tweets, *Kusuri* demonstrated performances close to human annotators with 93.7% F1-score, the best score achieved thus far on this corpus. On a corpus made of all tweets posted by 113 Twitter users (98,959 tweets, with only 0.26% mentioning medications), *Kusuri* obtained 76.3% F1-score. There is not a prior drug extraction system that compares running on such an extremely unbalanced dataset.

**Conclusion** The system identifies tweets mentioning drug names with performance high enough to ensure its usefulness and ready to be integrated in larger natural language processing systems.

*Keywords: Social Media, Pharmacovigilance, Drug Name Detection, Ensemble Learning, Text Classification*

# 1. Introduction

Twitter has been utilized as an important source of patient-generated data that can provide unique insights into population health [1]. Many of these studies involve retrieving tweets that mention drugs, for tasks such as syndromic surveillance [2] [3], pharmacovigilance [4], and monitoring drug abuse [5]. A common approach is to search for tweets containing a lexical match of drugs names occurring in a manually compiled dictionary. However, this approach has several limitations. Many tweets contain drugs that are misspelled or not referred to by name (e.g., '*it*' or '*antibiotic*'). Even when a match is found, oftentimes, the referent is not actually a drug; for example, tweets that mention *Lyrica* are predominantly about the singer, Lyrica Anderson, and not about the antiepileptic drug. In this study, when using the lexical match approach on a corpus where names of drugs are naturally rare, we retrieved only 67% of the tweets that we manually identified as mentioning a drug, and more than 47% of the tweets retrieved were noise. Enhancing the utility of social media for public health research requires methods that are capable of improving the detection of posts that mention drugs.

The task of automatically detecting mentions of concepts in text is generally referred to as Named Entity Recognition (NER) [6]. State-of-the-art NER systems are based on machine learning (ML) and achieve performances close to humans when they are trained and evaluated on formal texts. However, they tend to perform relatively poorly when they are trained and evaluated on social media [7]. Tweets are short messages, so they do not provide large contexts that NER systems can use to disambiguate concepts. Furthermore, the colloquial style of tweets—misspellings, elongations, abbreviations, neologisms, and non-standard grammatical structures, cases, and punctuations—poses challenges for computing features in ML-based NER systems. Although large sets of annotated corpora are available for training NER systems to detect general concepts on Twitter (e.g., people, organizations), automatically detecting more specialized concepts (e.g., drugs, diseases) requires collecting and annotating additional data [8]. Further complicating NER from social media, postings are usually collected over a short period of time. Given that comments on Twitter are often driven by news articles and have overrepresentation of certain NEs during a short period, and if this happens to coincide with the collection period, unexpected biases ensue, making them unsuitable to train a NER system [9]. For example, a corpus collected in November 2015 would have an abnormal frequency of tweets mentioning Daraprim due to the number of users commenting the large increase of the drug price, jumping from $13.50 a tablet to $750 overnight.

Over the last decade, researchers have competed to improve NERs on tweets. Most challenges were organized for tweets written in English, *e.g.* the Named Entity Recognition and Linking challenges series (NEEL) [10] or the Workshop on Noisy User-generated Text (W-NUT) series, but also in other languages such as Conference sur l'Apprentissage Automatique 2017 [11]. Specifically for drug detection in Twitter, we organized the 3[rd] Social Media Mining for Health Applications shared task in 2018. The sizes of the corpora annotated during these challenges vary, from 4000 tweets [12] to 10,000 tweets [11]. Technologies have changed over the years, with a noticeable shift from SVM or CRF machine learning frameworks trained on carefully engineered features [13] to Deep Neural Networks that automatically discover relevant features from word embeddings. In W-NUT'16, a large disparity between the performances obtained by the NERs on different types of entities was observed. The winners, [14], with an overall 52.4 F1-score, reported a 72.61 F-score on Geo-Locations, the most frequent NEs in their corpus, but much lower scores on rare NEs such as a 5.88 F1-score on TV Shows. *Kusuri* detects names of drugs with sufficient performance even on a natural corpus where drugs are very rarely mentioned.

The primary objective of this study is to automatically detect tweets that mention drug products (prescription and over-the-counter) and dietary supplements. The FDA[1] defines a drug product as the final form of a drug, containing the drug substance generally in association with other active or inactive ingredients. This study includes drug products that are referred to by their trademark names (e.g., *NyQuil*), generic names (e.g., acetaminophen), and class names (e.g., antibiotic). We formulate this problem as a binary classification task. A tweet is a positive example if it contains text referring to a drug (not only "matching" a drug name), a negative example otherwise. For example, the tweet "I didn't know Lyrica had a twin" is a negative example, Lyrica refers to the singer Anderson, whereas the tweet "Lyrica experiences? I was on Gabapentin." is a positive example, given the context, since it mentions two antiepileptics. We left as a future work the use of sequence labeling to delimit drug names boundaries in the positive examples.

The main contributions of this study are (1) a gold standard corpus of 15,005 annotated tweets, for training ML-based classifiers to automatically detect drug products mentioned on Twitter, (2) a binary classifier based on Ensemble Learning, which we call *Kusuri* ("medicine", or, more generally, "A substance that has a beneficial effect in improving or maintaining one's health" in Japanese[2]), and (3) an evaluation of *Kusuri* on 98,959 tweets with the natural balance of 0.2% positive to 99.8% negative for the presence of medication names. We describe the training and evaluation corpora in Section 3 as well as the details of our classifier followed by its evaluation in Section 4.

## 2. Related Work

Automatic drug name recognition has mostly been studied for extracting drug names from biomedical articles and medical documents, with several papers published [15] and challenges organized in the last decade [16] [17] [18].

Few works have tackled the task of detecting drug names in Twitter, most efforts were focused on building corpora. In [19], the authors created a large corpus of 260,000 tweets mentioning drugs. However, they restricted their search by strict matching to a preselected list of 250 drugs plus their variants, and they did not annotate the corpus generated. In a similar study, [20] explores the distributions of drug and disease names in Twitter as a preliminary work for Drug-Drug Interaction. While the authors searched for a larger set of drugs (all unambiguous drugs listed in DrugBank database), they did not annotate the corpus of 1.4 million tweets generated, and neither did they count False Positives (FP) in their statistical analysis.

The first evaluation of automatic drug name recognition in Twitter that we are aware of was performed by Jimeno-Yepes *et al.* in [21] on a corpus of 1,300 tweets. Two off-the-shelf classifiers, MetaMap and the Stanford NER tagger, as well as an in-house classifier based on a CRF with hand-crafted features were evaluated. The latter obtained the best F1-score of 65.8%. Aside from the afore mentioned problem of selecting the tweets using a lexical match, other limitations to their study lie in additional choices made. To remove non-medical tweets, they retained only tweets containing at least two medical concepts, *e.g.* drug and disease. This ensured a good precision but also artificially biased their corpus two ways: by retaining only the tweets that mentioned the drugs in their dictionary, and eliminating tweets that mention a drug alone, ex. *'me and ZzzQuil are best friends'*. In November 2018, we organized the 3rd Social Media Mining for Health Applications shared task (SMM4H) [22] with Task 1 of our challenge dedicated to the problem of the automatic recognition of drug names in Twitter. Eleven teams tested multiple approaches on a balanced corpus, selected using four classifiers (the first module of *Kusuri*). A wide range of deep learning-based classifiers were used by participants, but also feature-based classifiers and few attempts with ensemble learning systems. The system by Wu et *al.* [23], an ensemble of hierarchical neural networks with multi-head self-attention and integrating features modeling sentiments, was the top performer, with a 91.8% F1-score. This established a recent benchmark for the community for a balanced corpus (with approximately the same number of positive and negative examples). Our evaluation data, described in Section 3.2, includes both the automatically balanced corpus and, in addition, a corpus of all available tweets posted by selected Twitter users where the mention of drug products were manually annotated. We refer to the later as a "natural" corpus.

---

[1] FDA Glossary of Terms: https://www.fda.gov/drugs/informationondrugs/ucm079436.htm; Drug; Drug product; Accessed: 2018-05-18

[2] https://en.wiktionary.org/wiki/%E8%96%AC

## 3. Methods

We collected all publicly available tweets posted by 112,500 Twitter users (their *timelines*) [24]. To decide which users to include, we used a set of manually defined keywords and a simple classifier to detect tweets announcing a pregnancy. Once a tweet announcing a pregnancy was identified, we collected the timeline of the author of the tweet. We used the API provided by Twitter to download all tweets posted by this user within the Twitter-imposed limit of 3200 most recent tweets, and continued collection afterwards. Following this process, we collected a total of 421,5 million tweets. Using this corpus as a source allows us to avoid the bias of a drug-name keyword based collection.

All tweets were collected from public twitter accounts, and a certificate of exemption was obtained from the Institutional Review Board of the University of Pennsylvania. All tweets used and released to the community were used and released without violating Twitter's terms and conditions.

### 3.1 The UPennHLP Twitter Drug corpus

Building a corpus of tweets containing drug names to train and evaluate a drug name classifier is a challenging task. Tweets mentioning drug names are extremely rare. We found that they only represent 0.26% of the tweets in the *UPennHLP Twitter Pregnancy Corpus* (see next section), and are often ambiguous with common and proper nouns. If a naive lexicon matching method is used to create the corpus, it often matches a large number of tweets not containing any drug names.

Therefore, to build a gold-standard corpus we had to rely on a more sophisticated method than simply lexicon matching. We created four simple classifiers to detect tweets mentioning drug names, one based on a lexicon matching, one on lexical variants matching, one on regular expressions and a classifier trained with weak supervision. The four classifiers are described briefly below and in detail in Appendix A.

**Lexicon-based Drug Classifier** The first classifier is built on top of a lexicon of drugs names generated from the RxNorm Database[3]. If a tweet contains a word/phrase occurring in the lexicon, the tweet is classified as positive example without any further analysis. We chose RxNorm because it has a large coverage. It combines, in a unique database, 15 existing dictionaries, including DrugBank, a database often used in previous works on drug names detection.

**Variant-based Drug Classifier** Names of drugs may have a complex morphology and, as a consequence, are often misspelled on Twitter. Lexicon-based approaches detect drugs mentioned in tweets only if the drug names are correctly spelled. The incapability to detect misspelled drug names results in low recall for the lexicon-based classifier. In an attempt to increase recall, we used a data-centric misspelling

---

[3] RxNorm, https://www.nlm.nih.gov/research/umls/rxnorm/overview.html, Accessed: 2018-06-11

generation algorithm described in [25] to generate variants of drug names and used the variants to detect tweets mentioning misspelled drugs.

**Weakly-trained Drug LSTM Classifier** Our third classifier is a Long Short Term Memory (LSTM) neural network integrating an attention mechanism [26] and trained on noisy training examples obtained through weak supervision. One annotator identified drug names tending to be unambiguous [27] in our timelines (*e.g. Benadryl* or *Xanax*). We selected the ~126,500 tweets containing these unambiguous names in our timelines as positive examples. Then, given drug names occur very rarely in tweets, we randomly selected an additional ~126,500 tweets from our timelines as negative examples and trained our LSTM on these examples.

**Pattern-based Drug Classifier** Our last classifier implements a common method to detect general named entities, Regular Expressions (REs). REs describe precisely the linguistic contexts used in Twitter to speak about drugs. We manually crafted our REs by inspecting all N-Grams occurring before and after the most frequent unambiguous names of drugs in our ~126,500 tweets.

To obtain positive examples, we selected tweets retrieved by at least two classifiers, as they were most likely to mentioned drug names. To obtain negative examples, we selected tweets detected by only one classifier, given that if these tweets did not contain a drug name, they were non-obvious negative examples. Following this process, from our 421,5 million tweets we created a corpus of 15,005 tweets, henceforth referred to as the *UPennHLP Twitter Drug Corpus*. We removed from the corpus duplicated tweets, tweets not written in English and tweets that were no longer on Twitter (*e.g.* tweets deleted by the users) at the time of the collection. Two annotators familiar with the medical domain and social media annotated the corpus in its entirety, with a high Inter-Annotator Agreement (IAA) measured as Cohen's Kappa of 0.892. We randomly selected 9,623 tweets for training (4,975 positive and 4,648 negative examples) and 5,382 tweets for testing (2,852 positive and 2530 negative examples) and publicly released it to the research community for Task 1 of the SMM4H shared task [22].

### 3.2 UPennHLP Twitter Pregnancy corpus

A balanced corpus, such as the *UPennHLP Twitter Drug Corpus*, is a useful resource to study how people speak about drugs on social media. However, due to the mechanism of its construction, a balanced corpus does not represent the natural distribution of tweets mentioning drugs on Twitter. Consequently, any evaluation made on a balanced corpus will never be indicative of the performances to expect from a drug name classifier during a larger study. In order to further assess whether *Kusuri* could reliably be used in such a study, we ran additional experiments on the corpus of an epidemiologic study on birth defect outcomes in pregnancy [28]. For that, we collected 397 timelines of women tweeting during their pregnancy, and manually identified all drugs mentioned in the tweets posted during the period of pregnancy. It took 2.5 hours in average to annotate each timeline. We ran our experiments on a subset of 113 timelines (98,959 tweets) from this corpus[4], referred as *UPennHLP Twitter Pregnancy Corpus* in the remaining of the article.

### 3.3 Kusuri Architecture

*Kusuri*, described in Figure 1, applies sequentially two modules to detect the names of drugs in the *UPennHLP Twitter Pregnancy Corpus*. This section describes each module. The success of deep learning classifiers in natural language processing lies on their ability to discover automatically relevant linguistic features from word-embeddings [29] - an ability even more valuable when working on short and colloquial texts such as tweets. For this reason, we preferred to integrate in our modules deep learning classifiers over more traditional classifiers based on features engineering.

---
[4] Due to the large number of tweets annotated, we are still standardizing the annotations added by the 6 annotators into the corpus and no IAA was available at the time of writing.

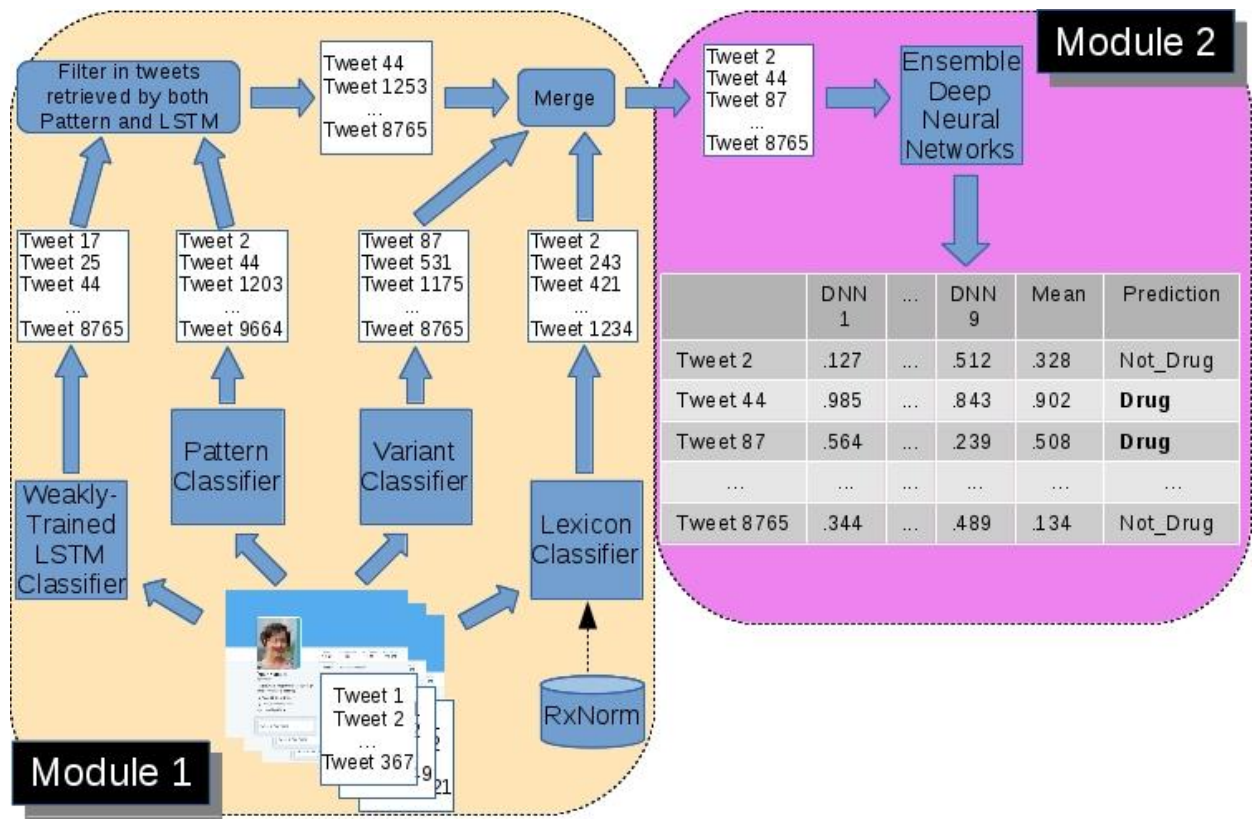

**Figure 1**: Architecture of Kusuri, an ensemble learning classifier for drug detection in Twitter

### 3.3.1 Module 1: Tweet Prefilter

*Kusuri* applies in parallel our four classifiers, the Lexicon-based, the variant-based, the pattern-based and the weakly trained LSTM classifiers, to discover tweets that potentially contain drug names. Among the tweets discovered, *Kusuri* selects the tweets classified by the Lexicon classifier, by the variant-based classifier and the tweets selected by both the pattern-based and the weakly trained classifiers. The tweets discovered by only one of the two last classifiers were too noisy and discarded. The tweets selected are then submitted to the Module 2, an ensemble of Deep Neural Networks (DNN) which takes the final decision for the labels. The four classifiers act as filters, collecting only good candidates for the ensemble of DNNs, which was, in turn, optimized to recognize positive examples among them.

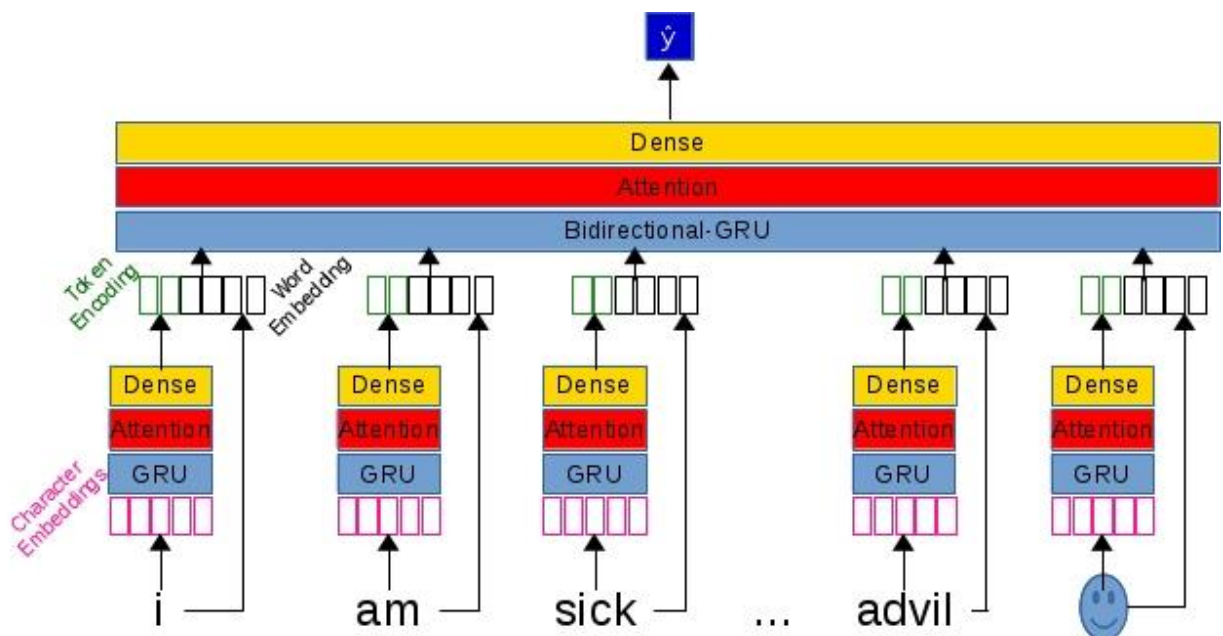

**Figure 2**: Deep Neural Network predicting ŷ, the probability for a tweet to mention a drug name

### 3.3.2 Module 2: Ensemble of Deep Neural Networks (DNN)

As single element for the ensemble of neural networks composing the second module of *Kusuri*, we designed a DNN following a standard architecture for classification of NEs. Described in Figure 2, our DNN starts by encoding independently each sequence of characters composing the tokens of a tweet through three layers sequentially connected, a recurrent layer, an attention layer and a densely connected layer. All resulting vectors, encoding the morphological properties of the tokens, are then concatenated with their respective pre-trained word embedding vectors, which encode the semantic properties of the tokens. The concatenated vectors are passed to a Bidirectional-GRU layer to learn long-range dependencies between the words, followed by an attention layer which, as additional memory, helps the NN to focus on the most differentiating words for the classification. A final dense layer computes the probability for the tweet to contain a mention of a drug. The Appendix B describes in detail the preprocessing steps, the embeddings and the parameters of our training.

Due to the stochastic nature of the initialization of the NN, the learning process may discover a local optimum and return a suboptimal model. To reduce the effect of local optimums, we resort to ensemble averaging, *i.e.*, we learned independently 9 models using our DNN and computed the final decision, for a tweet to mention a medication name or not, by taking the mean of the probabilities computed by the models[5]. When applied on the *Pregnancy Corpus*, all DNNs of the ensemble were trained on the *Drug Corpus* and, at test time, all DNNs of the ensemble only have to classify the tweets in the *Pregnancy Corpus* filtered by the first module of *Kusuri*.

Note that the ensemble of DNNs can be used to label tweets in any corpus. However, while the ensemble obtained state-of-the-art performances on the *Drug Corpus*, its performances dropped significantly when it labeled our *Pregnancy Corpus* without having the tweets prefiltered by the first module (discussed in Section 5).

---

[5] Since a soft voting algorithm [34] in our experiments did not improve over the simple averaging method, we kept the latter.

# 4. Results

This section details the performances of *Kusuri* and its ensemble of DNNs during two series of experiments on the *Drug Corpus* and on the *Pregnancy Corpus*.

## 4.1 Drug Detection in the UPennHLP Twitter Drug Corpus

We first ran a series of experiments to measure the performances of the ensemble of DNNs composing the second module of *Kusuri*. The detailed results are reported in Table 1. We compare the performance of our ensemble with three baseline classifiers.

The first baseline classifier, (1) in the table, is a combination of the Lexicon and variant-based classifier, it labeled as positive examples all tweets of the test set of the *Drug Corpus* that contained a phrase found in our Lexicon or in our list of variants. This baseline classifier provides a good estimation of the performances to expect when a Lexicon-based approach is used. We chose as a second baseline system a bidirectional-GRU classifier, (2) in the table. Since this baseline system has a simpler architecture than our final DNN, their comparison allows us to estimate the benefits of the components we added in our system. The baseline system was trained on the same training data with the same hyper-parameters and took as input the same word embeddings, but it did not have information about the morphology of the words neither the help of the attention mechanism. As a third strong baseline, we compare our system with the best system of the Task 1 of the SMM4H 2018 competition, the THU_NGN's system [23], (3) in the table.

| System | P | R | F1 |
| --- | --- | --- | --- |
| 1. Lexicon + Variant classifier | 66.4 | 88.5 | 75.9 |
| 2. Supervised Bidirectional-GRU | 93.5 | 89.5 | 91.4 |
| 3. THU_NGN's Hierarchical-NNs | 93.3 | 90.4 | 91.8 |
| 4. Best DNN model in the Ensemble | 93.7 | 92.5 | 93.1 |
| 5. **Ensemble DNNs (module 2 of Kusuri)** | **95.1** | **92.5** | **93.7** |

**Table 1:** Precision (P), Recall (R) and F1-Scores (F1) for Drug Detection Classifiers on the test set of the *UPennHLP Twitter Drug Corpus*

The results in Table 1 are interesting in several ways. The lexicon and variant-based classifier has a high recall on the test data with 88.5%, an unsurprising result considering the central role played by the lexicon and the variants during the construction of the *Drug Corpus*. This classifier is vulnerable to the frequent ambiguity of drug names, resulting in a low precision of 66.4%. The classifier has no knowledge of the context where a name of drug appears and cannot disambiguate such tweets, *e.g.* Lyrica (antiepileptic *vs* Lyrica Anderson), lozenge (type of pills *vs* geometric shape) or halls (brand name *vs* misspelling for Hall's). The fully supervised Bidirectional-GRU confirms its ability to learn the features only from the word embeddings [30] and achieves 91.2% F1-score, a higher score than the IAA computed on this corpus. However, such system can be improved as demonstrated by the better performances of the Best DNN in the ensemble DNNs, line (4) in Table 1. The encoding of the token morphology and the attention layer of the Best DNN are allowing an improvement of 1.7 F1-score points. Also, despite having a simpler architecture and attention mechanism, the Best DNN system performs better than the ensemble of hierarchical NNs proposed by [23]. The reason for this result is not clear but it may be a suboptimal set of hyper-parameters chosen by the authors or the difficulty to train such a complex network.

The highest performance is obtained by the ensemble DNNs which shows an improvement of 0.6 points over the best model in the ensemble with a final 93.7% F1-score[6]. We analyzed randomly selected labeling errors made by the Ensemble DNNs (Table 2). We distinguished eight, but non-exclusive, categories of FPs. With 41 cases, most FPs were tweets discussing medical topic without mentioning a drug. Since medical tweets often describe symptoms or discuss medical concepts, their lexical fields are strongly associated with drug names (*e.g.* cough, flu, doctor) and confuse the classifier. The causes of FNs seem to mirror those for the FPs. With 36 cases, FNs were mostly caused by the ambiguity, not only of common English words (*e.g.* airborne), but also of dietary supplements and food products sometimes consumed for their medicinal properties (*e.g.* clove, arnica or aloe). This could be a positive turn of events if nutritional supplements are to be included in a study. A second important cause was unseen, or rarely seen, drug names in our training corpus, with 25 cases.

| Error Category | #errors | Examples |
| --- | --- | --- |
| **False Positive** | | |
| Medical topic | 41 | <user> you should see a dermatologist if you can. You may just need something to break you out of a cycle. I used a topical and took pills <br> Lola may has a stye, or pink eye. Doc recommends warm compresses to see if it gets better today, but my eyes are itchy just looking at her. |
| Weighted words/patterns | 19 | *i can take a* wax brazillian a g <br> <user> i was robbed a foul when *i took a* three point shot and they got a few three pointers in. good game. |
| Ambiguous Name | 12 | <user> I actually really like *Lyrica* & A1. |
| Food topic | 11 | This aerobically fermented product was tested & it's antibiotic residue free. also certified organic. |
| Insufficient Context | 7 | <user> adding *Arnica* to my shopping list |
| Cosmetic topic | 5 | Doc prescribed me this dandruff shampoo, if it works, I'm definitely getting a sew in after I'm done using it |
| Unknown | 2 | Ice_Cream, Ice-Cream and More Ice-Cream...thats Ol i Want |
| Error annotation | 3 | --- |
| **False Negative** | | |
| Ambiguous Name | 36 | Trying Oil of Oregano & garlic for congestion for my sinus infection. [ambiguous dietary supplement] <br> In the church the person close to me's sniffling & coughing... I need a bathe of bactine and some *Airborne*, right now [ambiguous English word] |
| Drug not/rarely seen | 25 | That's the *benzo* effects! [missing variant] <br> Pennsylvania Appellate Court Revives 1,000 *Prempro* Cases Against Pfizer [missing in lexicon] <br> the *percocet-thief* plot makes Real World New Orleans look almost intriguing [preprocessing error] |
| Generic terms | 18 | Tossing and turning. I need ur *sleep aid*. Waiting patiently <user> |
| Non-medical context | 11 | I have a love/hate relationship with retin A. #amusthave. |
| Short tweets | 3 | arnica-ointment-7 |
| Error annotation | 7 | --- |

**Table 2**: Categories of False Positive and False Negative made by the Drug Detection Classifier on the test set of the *UPennHLP Twitter Drug Corpus*

---

[6] We confirmed the disagreement between the ensemble DNNs and the THU_NGN systems to be statistically significant with a McNemar's test [35], the null hypothesis was rejected with a significance level set to .001

## 4.2 Drug Detection in the UPennHLP Twitter Pregnancy Corpus

The results of the ensemble DNNs on the *Drug Corpus* are good but obtained on ideal conditions. The training corpus is balanced and most of the drugs found in the test set were present in the training set. These conditions are unlikely to be satisfied when the classifier is used on naturally occurring data. We ran a second series of experiments on the *Pregnancy Corpus* to get a better evaluation of our classifiers. As for the previous experiments, we kept the lexicon and variant-based classifier as a baseline system. Since the ensemble DNNs gave the best performances on the *Drug Corpus* (Table 1), we chose it as a second baseline system (i.e. this baseline applies the ensemble of DNNs without prefiltering the tweets using the first module of *Kusuri*). The last system evaluated was the "complete" *Kusuri* system, with both modules applied sequentially.

| System | P | R | F1 |
|---|---|---|---|
| 1. Lexicon + Variant classifier | 53.4 | 67.1 | 59.5 |
| 2. Ensemble DNNs (Only module 2 –classifier- of Kusuri) | 10.5 | 84.1 | 18.7 |
| 3. Kusuri (module 1 –filters- + module 2 –classifier-) | **95.3** | 63.6 | **76.3** |

**Table 3:** Precision (P), Recall (R) and F1-scores (F1) for Drug Detection Classifiers on the *UPenn Twitter Pregnancy Corpus*

The results are reported in Table 3. What is striking about the figures in this table is the poor performances of the lexicon/variant-based and the ensemble DNNs classifiers. The drop in the score of the former, 59.5% F1-score against 75.9% F1-score on the *Drug Corpus*, is a reflection on the "beneficial" effect of the bias introduced when building the corpus. The drugs in the lexicon were over represented, increasing its recall by 21.4 points. The ensemble DNNs classifier did worse with only 18.7% F1-score. Trained on a balanced set of medically related tweets, the classifier was found too sensitive. It gives too much weights to words related to medication but used in other contexts such as *overdose*, *bi-polar* or *isnt working*, resulting in a total of 1,844 FPs, where only 258 tweets mentioned a drug in the timelines. While lower than the ideal score of the ensemble DNNs classifier on the *Drug Corpus*, the score of *Kusuri*, 76.3 F1-score, is comparable to the scores published for the best NERs when applied on the most frequent types of NEs in Twitter [14]. More importantly, we believe this score is high enough to expect a positive impact of *Kusuri* when integrated in larger applications.

## 5. Conclusion and Future Work

In this paper, we presented *Kusuri*, an ensemble learning classifier to identify tweets mentioning drug names. Given the unavailability of a labeled corpus to train our system, we created and annotated a balanced corpus of 15,005 tweets. The ensemble of deep neural networks at the core of *Kusuri*'s decisions (Module 2), with 93.7% F1-score demonstrated performances close to human annotators without requiring engineered features on this corpus. However, since we built this corpus artificially, it did not represent the natural distribution of drug mentions in Twitter. We evaluated *Kusuri* on a second corpus made of all tweets posted by 113 Twitter users, a total of 98,959 annotated tweets with only 258 tweets mentioning drugs. On this corpus, *Kusuri* obtained 76.3% F1-score, a score comparable to the score obtained on the most frequent types of NEs by the best systems competing in well-established challenges, despite our corpus having only 0.26% positive instances in it.

Several limitations can be addressed to potentially improve the ensemble DNNs' performances on the balanced corpus. We designed our neural network around a standard representation of a sentence as a sequence of word embeddings learned on local windows of words. Better alternatives have been recently proposed [31] and could be integrated in our system to help drug names disambiguation. Such changes could be replacing our word embeddings with ELMo, which learns each word embeddings within the whole context of a sentence [32], or complementing our current sentence representation with a sentence embeddings [33].

However, the main challenge comes from the extreme difference in the negative to positive ratio in the natural corpus. Our strategy reduces the number of FPs with hard filters and consequently, also limits the overall performance of *Kusuri* by removing 32% of the few drug names mentioned in the corpus. We are currently replacing the hard filters with active learning to further train our ensemble of DNNs and reduce their over-sensitivity to medical phrases in general tweets.

## Funding

Research reported in this publication was supported by the National Library of Medicine (NLM) under grant number R01LM011176. The content is solely the responsibility of the authors and does not necessarily represent the official view of NLM.

*Conflict of Interest:* none declared